\pdfoutput=1

\documentclass[11pt]{article}

\usepackage[final]{acl}

\usepackage{times}
\usepackage{latexsym}
\usepackage{multirow}

\usepackage[T1]{fontenc}

\usepackage[utf8]{inputenc}

\usepackage{microtype}

\usepackage{inconsolata}

\usepackage{graphicx}
\usepackage{booktabs}
\usepackage{amsmath}
\usepackage{amssymb}
\definecolor{babyblueeyes}{rgb}{0.63, 0.79, 0.95}
\definecolor{antiquewhite}{rgb}{0.98, 0.92, 0.84}
\usepackage{caption}
\usepackage{footnote}
\makesavenoteenv{figure}

\title{StyEmp: Stylizing Empathetic Response Generation via \\
Multi-Grained Prefix Encoder and Personality Reinforcement
}

\author{Yahui Fu, Chenhui Chu, and Tatsuya Kawahara \\ Graduate School of Informatics, Kyoto University, Japan\\ \texttt{[fu, kawahara]@sap.ist.i.kyoto-u.ac.jp}\\
\texttt{chu@i.kyoto-u.ac.jp}}

\begin{document}
\maketitle\begin{abstract}
Recent approaches for empathetic response generation mainly focus on emotional resonance and user understanding, without considering the system's personality. Consistent personality is evident in real human expression and is important for creating trustworthy systems. To address this problem, we propose StyEmp, which aims to stylize the empathetic response generation with a consistent personality. Specifically, it incorporates a multi-grained prefix mechanism designed to capture the intricate relationship between a system's personality and its empathetic expressions. Furthermore, we introduce a personality reinforcement module that leverages contrastive learning to calibrate the generation model, ensuring that responses are both empathetic and reflective of a distinct personality. Automatic and human evaluations on the EMPATHETICDIALOGUES benchmark show that StyEmp outperforms competitive baselines in terms of both empathy and personality expressions.\footnote{Our source code is publicly available at  \href{https://github.com/fuyahuii/StyEmp}{https://github.com/fuyahuii/StyEmp}.}
\end{abstract}

\section{Introduction}
Empathy and personality are pivotal factors in the development of human-like systems. Empathy is the ability of humans to put themselves in another's position, which encompasses understanding another's experiences and feelings for responding appropriately. Personality is the enduring patterns of thoughts, feelings, and behaviors that distinguish individuals from one another \cite{allport1937personality}.

\begin{figure}
    \centering
\includegraphics[width=0.49\textwidth]{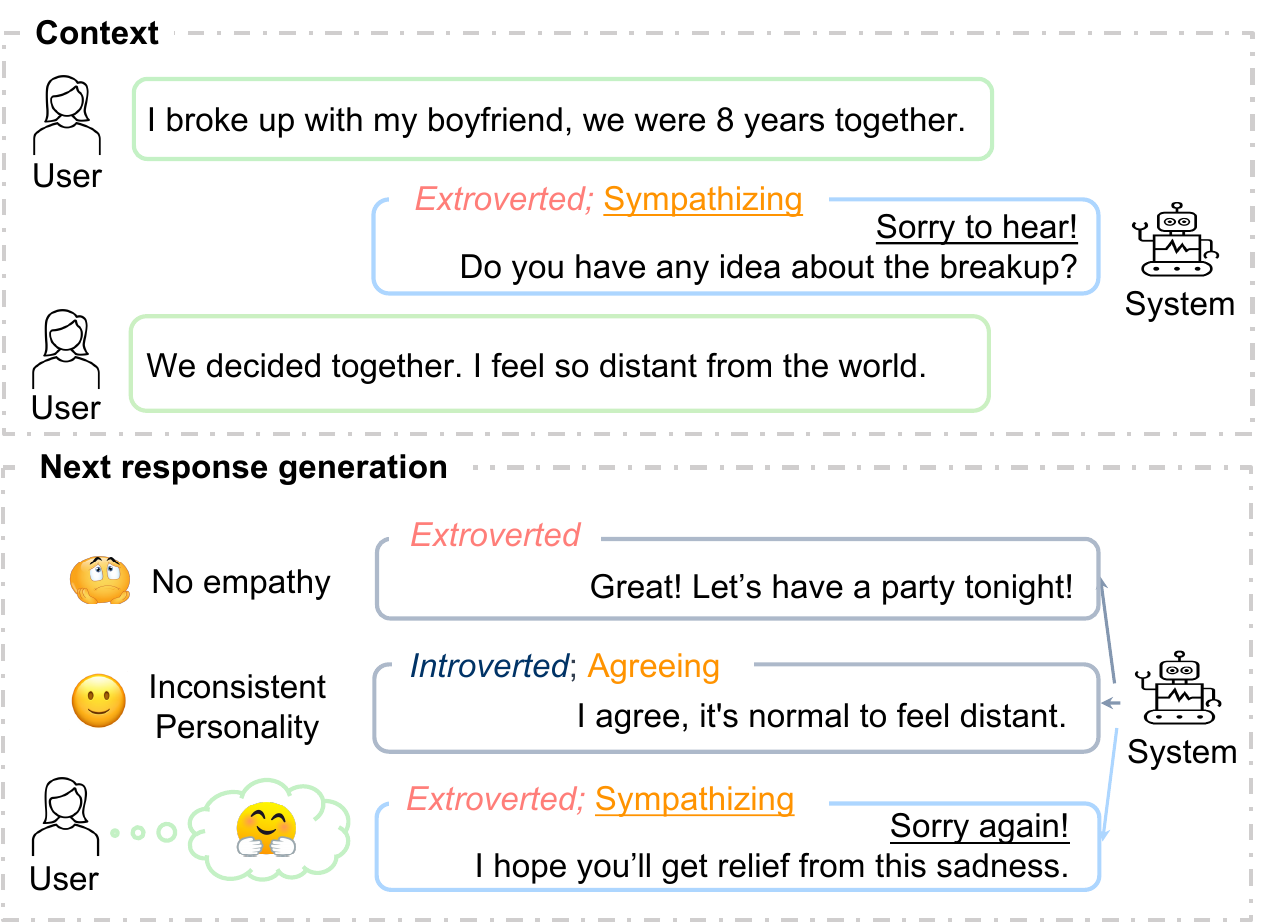}
        \caption{Different \textit{personalities} exhibit distinct preferences for 
        \textcolor{orange}{empathetic intents\protect\footnote{We utilize nine empathetic intents from \citet{welivita2020taxonomy}, which do not strictly adhere to the definition of empathetic, including sympathizing and agreeing.}}
        in responses \cite{richendoller1994exploring, mairesse2010towards}.
        In a given context, the user shows varying feelings to the system's responses, where the system encompasses empathetic expression and consistent personality traits, resulting in a more human-like interaction.} 
    \label{fig:example}
\end{figure}

Empathy integrates cognition and emotion, involving understanding and responding emotionally to others' situations \cite{davis1983measuring}. Consequently, prior research has focused on methods to generate empathetic responses by improving affective expression \cite{lin2019moel, majumder2020mime, li2020empdg}, or exploring context understanding \cite{majumder2022exemplars, wang2022care,sabour2022cem, fu2023reasoning}. 
However, as illustrated in Figure~\ref{fig:example}, individuals with different personalities can exhibit diverse empathy styles given identical contexts. Previous methods for empathetic response generation did not consider the system's personalities, which leads to responses that may reflect empathy but lack personalization.

Systems that express a consistent personality are important for enhancing believability \cite{higashinaka2018role}. As shown in Figure~\ref{fig:example}, when the system changes its personality in a conversation, it would make the interaction feel less human-like. 
Moreover, an appropriate empathetic response may depend on the personality traits.
\citet{richendoller1994exploring} 
examined the relationships between psychoticism, extraversion, and neuroticism and three styles of empathic intents: empathetic, perspective-taking, and sympathetic. Their findings indicate that individuals with different personalities exhibit distinct preferences for empathetic intents, inspiring our motivation to consider the system's personality traits in empathetic response generation. However, the relationship between commonly-used Big 5 \cite{mccrae1992introduction} / Myers-Briggs Type Indicator (MBTI) \cite{myers1962myers} personalities and empathetic intents has not been fully explored.

To address this, we implicitly learn these connections through the prediction of both personality traits and empathetic signals in responses. Empathetic signals include empathetic intents and empathetic communication mechanisms (ECM) \cite{sharma2021computational} - interpretations (IP), explorations (EX), and emotional reactions (ER). Further inspired by the prefix tuning method employed by \citet{li-liang-2021-prefix} and \citet{liu-etal-2023-recap}, we propose a multi-grained prefix encoder aimed at discerning personality traits alongside empathetic signals.

Because the EMPATHETICDIALOGUES dataset (ED) \cite{rashkin2019towards} primarily targets expressing empathy rather than personality, it is hard to learn personality traits from a single response. To solve this problem, we utilize a pool of past utterances by the same listener to predict and encode personality traits. Then, we propose a personality reinforcement (PR) module to calibrate the generation of empathetic responses by integrating explicitly personality traits. 
Our main contributions are: 
\begin{itemize}
    \item To the best of our knowledge, this is the first work to consider the system's personality for empathetic response generation. Moreover, we propose a multi-grained prefix mechanism to implicitly learn the relationship between the system's personality and corresponding empathetic expressions.
    \item We introduce a personality reinforcement module to calibrate an empathetic response generation model via contrastive learning for generating responses that are both empathetic and reflective of a distinct personality.
\end{itemize}

\section{Related Work}
\subsection{Empathetic Response Generation}
Previous approaches to empathetic response generation mainly align with three categories: The first category emphasizes the affective aspect of emotional expression, detecting and leveraging the user's emotion using various structures \cite{lin2019moel, majumder2020mime, li2020empdg}. The second category focuses on contextual understanding through different mechanisms, including 
emotion cause reasoning \cite{kim2021perspective, wang2022care}, additional retrieval processes \cite{majumder2022exemplars, fu2023dual}, and integration of commonsense knowledge \cite{li2022knowledge,sabour2022cem,fu2023reasoning}. The third category augments large language models (LLMs)'s capabilities in empathetic expression \cite{lee2022does, zhao2023chatgpt}.
However, these methods often ignore the personality traits evident in empathetic expressions, leading to responses that exhibit inconsistent personalities. To address this discrepancy, our study predicts both personality traits and empathetic signals, introducing a multi-grained prefix encoder designed to implicitly learn the connections between them.


\subsection{Personalized Response Generation}
Recent advancements in personalized response generation fall into three distinct categories: (1) generation based on explicit personality traits, such as those characterized by the Big 5 model \cite{saha2022stylistic,xu2023generating,ramirez2023controlling}. (2) customization using explicit system-specific profiles or descriptive persona sentences \cite{zhang2018personalizing, mazare2018training,zhong2020towards}. (3) tailoring responses according to an implicit system persona derived from past responses \cite{zhong2022less, liu-etal-2023-recap}. Manual collection of explicit system personalities or persona profiles is both time-consuming and costly. To avoid it, we learn the implicit system's personality from their past responses and incorporate explicit personality expression through an additional personality reinforcement module via contrastive learning.


\section{Preliminaries}
Due to the lack of personality and empathetic signal annotations within the benchmark ED dataset, we train distinct models specialized for each aspect. 

\subsection{Personality Predictor}


\begin{table}[htp]
\centering
\scalebox{0.81}{
\begin{tabular}{@{}lccccc@{}}
\toprule
     Traits &Acc. & BA. & F1  & Pear. & Spear. \\
                             \midrule
  \textbf{Introverted}            & \textbf{59.11}             & \textbf{58.15}    & \textbf{65.41}                   & \textbf{\textit{0.1838}}  & \textbf{\textit{0.1852}}   \\
                       Intuitive               & 50.50             & 50.39      & 56.83                 & \textit{-0.0592} & \textit{-0.0506}  \\
                       \textbf{Thinking}                & \textbf{59.30}         & \textbf{59.06}          & 55.79                 & \textbf{\textit{0.2344}}  & \textbf{\textit{0.2287}}   \\
                       Perceiving              & 49.16           & 49.26        & 47.00                 & \textit{-0.0166} & \textit{-0.0157}  \\ \midrule
  Agreeable                     & 47.72           & 47.45        & 0.5468         & \textit{-0.0274} & \textit{-0.0312}  \\
                       Conscientious                    & 52.46        & 53.75          & 0.5663                 & \textit{0.1291}  & \textit{0.1016}   \\
                       \textbf{Extraversion}                     & \textbf{67.23}        & \textbf{63.70}           & \textbf{0.7566}                & \textbf{\textit{0.4081}}  & \textbf{\textit{0.3862}}   \\
                       Neuroticism                     & 53.91         & 54.02         & 0.5696                 & \textit{0.1074}  & \textit{0.1025}   \\
                       Openness                     & 50.06     & 49.88    & 0.5338   & \textit{0.0466}  & \textit{0.0511}   \\ \bottomrule
\end{tabular}}
\caption{Accuracy and correlation results of MBTI and Big 5 based on the Pandora dataset. Pear. and Spear. denote the Pearson/Spearman correlation between prediction and ground truth on each personality trait, \textit{Italics} mean statistical significant ($p<.05$).}
\label{tab:personalityacc}
\end{table}

PANDORA \cite{gjurkovic-etal-2021-pandora}\footnote{\href{{https://psy.takelab.fer.hr/datasets/all/pandora}}{https://psy.takelab.fer.hr/datasets/all/pandora}} is the largest dataset of Reddit comments labeled with Big 5 and MBTI traits intensities. 
We strictly partition the PANDORA dataset by the user, guaranteeing no user overlap across the training, validation, and test sets. This approach allows us to assess the model's efficacy in identifying the personality traits of unseen users, thereby making the evaluation results on the PANDORA dataset applicable to the ED dataset as well.
We finetune LUKE \cite{yamada-etal-2020-luke}\footnote{\href{https://huggingface.co/studio-ousia/luke-base}{https://huggingface.co/studio-ousia/luke-base}} model with regression head for automatically detecting Big 5 and MBTI personality traits using the PANDORA dataset.
Based on the prediction accuracy shown in Table~\ref{tab:personalityacc}, we adopt the combination of MBTI introverted, MBTI thinking, and Big 5 extraversion as personality traits used in this study. More experimental details and results can be seen in Appendix~\ref{sec:personality_anno}.

\subsection{ECM and Intent Predictor}
Empathetic signals comprise both ECM and intent, which are complementary. For example, \textit{Encouraging} or \textit{Sympathizing} in intent prediction is detailed beyond \textit{Interpretation} in the ECM. Additionally, ER within the ECM dictates whether a response contains emotional signals.

\noindent\textbf{ECM}: Inspired by \citet{lee2022does, fu2023reasoning, bi-etal-2023-diffusemp}, 
we use \textit{IP}, \textit{EX}, \textit{ER} as parts of the empathetic signals. Specifically, \textit{IP} represents expressions of acknowledgments
or understanding of the interlocutor’s emotion or
situation. \textit{EX} represents expressions of active interest
in the interlocutor’s situation; \textit{ER} represents expressions of explicit emotions. Specifically, we follow official codes\footnote{\href{https://github.com/behavioral-data/Empathy-Mental-Health}{https://github.com/behavioral-data/Empathy-Mental-Health}} and use three RoBERTa-based \cite{liu2019roberta} classifiers to identify whether a response implies a certain trait individually.

\begin{table}[t]
\centering
\begin{tabular}{@{}cr|rrr@{}}
\toprule
Traits  &   \#Classes &Acc.    & BA. & F1    \\ 
\midrule
ER & 2&  84.76 & 84.13          & 84.70        \\
IP  &2   & 84.12 & 85.35          & 84.23       \\
EX   &2     & 94.81 & 92.46          & 94.86       \\
EI  &9& 90.17 & 90.17          & 90.23 \\
\bottomrule
\end{tabular}
\caption{Evaluations on empathetic signals predictor. ER, IP, EX, and EI denote Emotional Reaction, Interpretation, Exploration, Empathetic Intent classification, respectively. Acc. and BA. denote accuracy and balanced accuracy, respectively.}
\label{tab:empathyaccu}
\end{table}

\noindent\textbf{Intent}: 
Prior research by \citet{welivita2020taxonomy} highlighted incorporating dialogue intent modeling into response generation enhances the controllability and interpretability of generated responses. Then they introduced the EmpatheticIntents dataset,\footnote{\href{https://github.com/anuradha1992/EmpatheticIntents}{https://github.com/anuradha1992/EmpatheticIntents}} which is enriched with intent annotations, such as \textit{Suggesting}, \textit{Acknowledging}, and \textit{Agreeing}. We finetune a RoBERTa-base \cite{liu2019roberta} model on nine-class intent classification to label responses.
The results are shown in Table~\ref{tab:empathyaccu}.

\begin{figure*}
    \centering
\includegraphics[width=0.85\textwidth]{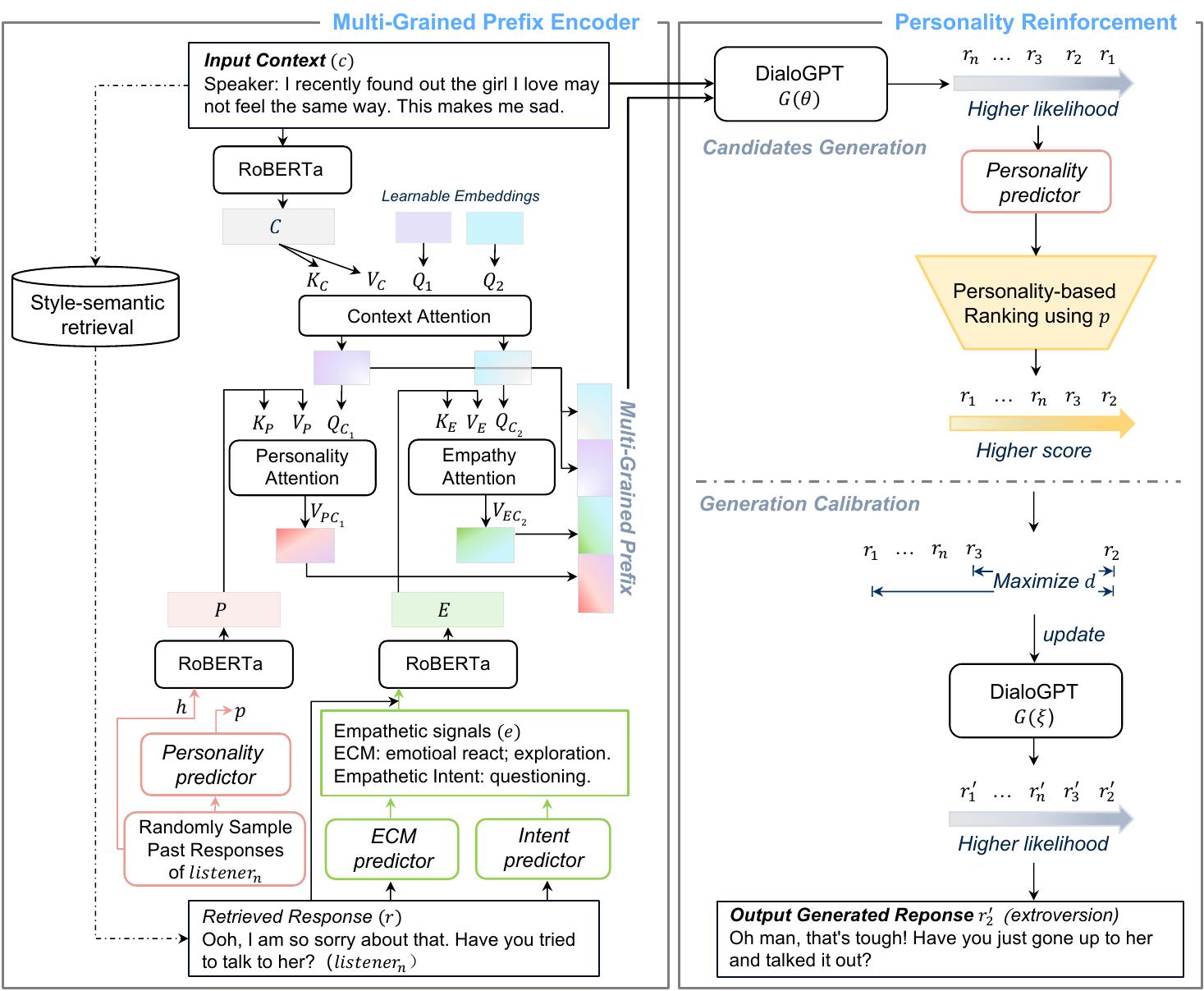}
    \caption{The architecture of our proposed method that contains a multi-grained prefix encoder and personality reinforcement module.}
    \label{fig:arch}
\end{figure*}

\section{Proposed Method}
Figure~\ref{fig:arch} shows an overview of our proposed method which comprises two main components. 
Firstly, a multi-grained prefix encoder is designed to implicitly learn the connections between personality traits and empathetic signals present in the system's response by multi-grained signals prediction and prefix encoding.
Secondly, we introduce a personality reinforcement mechanism aiming at integrating the generation of empathetic responses with explicit personality trait learning. 

\subsection{Mutli-Grained Prefix Encoder}
There are 810 unique listeners in the benchmark ED dataset, and each participant is involved in up to 100 conversations. Based on the listener ID, we sampled ten past responses by the same listener from the training set to implicitly learn listener's personality.
Inspired by the prefix-tuning mechanism employed in \citet{li-liang-2021-prefix}, \citet{liu-etal-2022-p}, and \citet{liu-etal-2023-recap}, we project the input context ($c$), the concatenation of retrieved response ($r$) (refer to Section~\ref{section:training and inference}) and empathy signals ($e$), and listener's past responses ($h$) into fixed-length prefix vectors, which are then prepended to the decoder hidden states as a prefix. 

We first use the RoBERTa model to encode the $c$, $e$ and $h$ to continuous representations, denoted as $\mathbf{C}$, $\mathbf{P}$, $\mathbf{E}$:
\begin{align}
    \mathbf{C}&=\text{RoBERTa}(c)\\
    \mathbf{P}&=\text{RoBERTa}(h)\\
    \mathbf{E}&=\text{RoBERTa}(\text{concat}(r,e))
\end{align}
To separately extract distinct context-related empathy and personality features, we introduce two learnable embeddings to act as distinct queries, $\mathbf{Q_{1}}$ and $\mathbf{Q_{2}}$, where $\mathbf{Q_{1}}$ is in $\mathbb{R}^{dn_{1}}$ and $\mathbf{Q_{2}}$ in $\mathbb{R}^{dn_{2}}$; here, $d$ represents the dimension of the RoBERT's last hidden layer, while $n_{1}$ and $n_{2}$ denote the lengths of the respective queries. The context representation $\mathbf{C}$, serves as both key $\mathbf{K_{C}}$ and value $\mathbf{V_{C}}$. Employing a cross-attention mechanism, we project context $\mathbf{C}$ into two fixed-length prefix vectors. These vectors are subsequently treated as $\mathbf{Q_{C_{1}}}$ and $\mathbf{Q_{C_{2}}}$: 
\begin{align}
    \mathbf{Q_{C_{1}}}=\text{Attn}(\mathbf{K_{C}}, \mathbf{V_{C}},\mathbf{Q_{1}})\\
    \mathbf{Q_{C_{2}}}=\text{Attn}(\mathbf{K_{C}}, \mathbf{V_{C}},\mathbf{Q_{2}})
\end{align}
Then following the same process, we fuse the representations of the listener's past responses $\mathbf{P}$, and the empathy explanation representations $\mathbf{E}$, with the context-related prefix vectors $\mathbf{Q_{C_{1}}}$ and $\mathbf{Q_{C_{2}}}$, respectively: 
\begin{align}
    \mathbf{V_{PC_{1}}}&=\text{Attn}(\mathbf{K_{P}}, \mathbf{V_{{P}}},\mathbf{Q_{C_{1}}})\\
    \mathbf{V_{EC_{2}}}&=\text{Attn}(\mathbf{K_{E}}, \mathbf{V_{E}},\mathbf{Q_{C_{2}}})
\end{align}
This fusion process yields two distinct vectors: $\mathbf{V_{PC_{1}}}$, which encapsulates the context-personality relationship, and $\mathbf{V_{EC_{2}}}$, representing the context-empathy relationship. This ensures that both personality and empathy dimensions are considered in the context of the interaction.

We then concatenate $\mathbf{Q_{C_{1}}}$, $\mathbf{Q_{C_{2}}}$, $\mathbf{V_{PC_{1}}}$, and $\mathbf{V_{EC_{2}}}$ by the length dimension, followed by one linear layer, to produce the final representations $\mathbb{R}^{2(n_{1}+n_{2})*d}$, as the final prefix embeddings.

\subsection{Decoder}
We utilize the pretrained DialoGPT \cite{zhang-etal-2020-dialogpt}\footnote{\href{https://huggingface.co/docs/transformers/model\_doc/dialogpt}{https://huggingface.co/docs/transformers/model-doc/dialogpt}} as the decoder. 
We further feed the final prefix embeddings into DialoGPT-small and train the parameters in the model on the ED dataset, then obtain a base empathetic response generator $G({\theta})$.

\subsection{Personality Reinforcement}
Because the ED dataset primarily targets expressing empathy rather than personality, it is hard to learn personality traits from a single response with traditional backpropagation. 
Drawing inspiration from recent calibration work \cite{zhang2022momentum, liu-etal-2022-brio, jiashuo2023aligning}, we generate multiple candidate responses via diverse beam search \cite{vijayakumar2016diverse}, which exhibit similar levels of empathy but vary in the degree of personality expressed. Subsequently, the proposed personality-based ranking module evaluates and ranks these candidates. Then, we calibrate the generation process by integrating a personality-oriented contrastive loss alongside the empathy loss, thereby achieving a generation of empathetic responses that reflect explicit personality traits.

\subsubsection{Candidate Generation}
For a input context $c$, we use the trained model $G({\theta})$ to generate $K$ empathetic candidate responses by diverse beam search: {${r_{1}, r_{2}, r_{3},...,r_{K}}$}, which can encapsulate varying degrees of personality expression.

\subsubsection{Personality-based Ranking}
We utilize our pretrained personality predictor, which estimates the system's personality $p$ from the past responses ($h$), including Big 5 extroversion ($p_{e}$), MBTI introversion ($p_{i}$), and MBTI thinking ($p_{t}$). Then, we predict the personality traits of each candidate in \(\{r_1, r_2, r_3, \ldots, r_K\}\), and calculate their personality margin \(S_{r_k}\). This margin is derived as the sum of the mean square errors (MSE) between the personality scores \(p\) and the predicted scores for each trait, formulated as:
\begin{equation}
    S_{r_k} =\left | p_{e}'-p_{e} \right |^{2}+\left | p_{i}'-p_{i} \right |^{2}+\left | p_{t}'-p_{t} \right |^{2}
\end{equation}
where $p_{e}'$, $p_{i}'$, and $p_{t}'$ are the predicted scores for each candidate on extroversion, introversion, and thinking traits, respectively. Based on this personality margins, we re-rank all candidate responses in ascending order of \(S_{r_k}\):  \(\{r_1^{'}, r_2^{'}, \ldots, r_K^{'}\}\),
where \(S_{r_i^{'}} < S_{r_j^{'}} \), for \(\forall i < j\).

\subsubsection{Generation Calibration}
We aim to encourage the model to assign higher estimated probabilities to empathetic candidate response with lower personality margin by adjusting the model $G({\theta})$ with a contrastive loss. Following the previous work \cite{zhang2022momentum,liu-etal-2022-brio,jiashuo2023aligning}, the pairwise margin loss is defined as:
\begin{gather}
\mathcal{L}_{p}=\sum_{i}\sum_{j>i}\mathrm{max}(0,p(r_{j}^{'}|c;\xi)-p(r_{i}^{'}|c;\xi)+\lambda_{i,j})
\end{gather}
where $\lambda_{i,j}$ is the dynamic margin multiplied by the difference in rank between the candidates,  $\lambda_{i,j}= \alpha \ast (j-i)$, and  $\alpha$ is a hyper-parameter. $p(r_{i}^{'}|c; \xi)$ is the generation probability computed by DialoGPT.

\subsection{Training and Inference}
\label{section:training and inference}
\noindent\textbf{Training} During the training phase, we use the ground truth as the retrieved response for empathy and intent prediction, and randomly sample the past responses of the corresponding listener. We aim to generate responses that are both good at empathy and personality expression,
then the final negative log-likelihood for generation is defined as:
\begin{equation}
    \mathcal{L}= -\sum\nolimits_{t=1}^{|y|}\log p\left(y_{t}|c, y_{<t};\xi\right)+\beta \mathcal{L}_{p}
\label{eq:training}
\end{equation}
where $\beta$ are hyper-parameters to balance the empathy and personality loss. We minimize $\mathcal{L}$ to optimize the generator's parameters $\xi$. 

\noindent\textbf{Inference}
During the inference phase, 
we employ a style-semantic retrieval mechanism that matches each test-set context (input) with similar contexts in the training set. The most similar context's corresponding response is treated as the retrieved response. Based on the listener ID associated with this response, we sample past responses. Considering the importance of emotion, semantics, and style in empathy and personality expression, we focus on these dimensions during the retrieval process.
Specifically, we utilize Sentence-BERT \cite{reimers2019sentence}\footnote {\href{https://huggingface.co/sentence-transformers/all-MiniLM-L6-v2}{https://huggingface.co/sentence-transformers/all-MiniLM-L6-v2}} to obtain semantic embeddings. We employ an off-the-shelf, content-independent style representation model \cite{wegmann2022same}\footnote{\href{https://huggingface.co/AnnaWegmann/Style-Embedding}{https://huggingface.co/AnnaWegmann/Style-Embedding}} for style embeddings. Furthermore, to enhance emotional relevance, we finetune RoBERTa \cite{liu2019roberta}\footnote{\href{https://huggingface.co/FacebookAI/roberta-base}{https://huggingface.co/FacebookAI/roberta-base}} on the ED dataset, targeting a classification of 32 emotions, the accuracy of which is 56.06\%.
Subsequently, we extract emotional embeddings from the final layer of the finetuned RoBERTa model. 
The final retrieval score is:
\begin{equation}
\mathrm{score}=\mathrm{sim}_{sem}+\mathrm{sim}_{style}+\mathrm{sim}_{emo}
\label{eq:sim}
\end{equation}
where $\mathrm{sim}_{sem}$, $\mathrm{sim}_{style}$, and $\mathrm{sim}_{emo}$ represent similarity in semantics, style, and emotion, respectively.

\section{Experimental Settings}
\subsection{Dataset}
The EMPATHETICDIALOGUES dataset \cite{rashkin2019towards}\footnote{\href{https://huggingface.co/datasets/empathetic\_dialogues}{https://huggingface.co/datasets/empathetic\_dialogues}} comprises 25k open-domain multi-turn conversations between two interlocutors. 
We train and evaluate our model for each turn of \textit{Listener} responding to \textit{Speaker}, and extend \textit{Speaker}'s inquiries one by one from the context history. The ratio for training/validation/test is roughly 8:1:1.

\subsection{Settings}
Our implementation is based on Huggingface's Transformers.\footnote{\href{https://huggingface.co/docs/transformers}{https://huggingface.co/docs/transformers}} For the multi-grained prefix encoder, we train Roberta as an encoder and DialoGPT-small as a decoder from scratch on the ED dataset. We set the learning rate to 5e-5, and batch size to 64. In the encoder configuration, the query length is set to 30. 
We sample 10 past responses by the same listener from the training set.
In the decoder configuration, the number of candidates $K$ is set to 5.
For the personality reinforcement, we set $\alpha$ and $\beta$ to be 0.001 and 1, respectively.
For the response generator, we use nucleus sampling (top-$p$) \cite{holtzman2019curious} with $p$ set to 0.8 and temperature to 0.7. All experiments use the same seed to minimize the impact of randomness.

\subsection{Models}
\subsubsection{Comparative Baselines}
\noindent \textbf{Transformer-based methods}
\footnote{\href{https://github.com/Sahandfer/CEM} {https://github.com/Sahandfer/CEM}}:

\noindent \textbf{MoEL} \cite{lin2019moel}: which softly combines multiple emotion-specific decoders to a meta decoder to generate empathetic responses.

\noindent \textbf{MIME} \cite{majumder2020mime}: integrates emotion grouping, emotion mimicry, and stochasticity into the mixture for various empathetic responses.

\noindent \textbf{EmpDG} \cite{li2020empdg}: which learns emotions and responses based on adversarial learning. 

\noindent \textbf{CEM} \cite{sabour2022cem}: which employs commonsense knowledge, 
to enhance its understanding of the interlocutor's situations and emotions.

\noindent \textbf{Large language model (LLM)-based methods}:
\noindent \textbf{DialoGPT} \cite{zhang-etal-2020-dialogpt}: a GPT2 model trained on Reddit conversation, we finetune it on the ED dataset for empathetic response generation.

\noindent \textbf{LEMPEx}\cite{majumder2022exemplars}: adopts T5 as the encoder-decoder and utilizes a combination of exemplar-based retrieval, a generator, and an empathy control module for empathy generation.\footnote{\href{https://github.com/declare-lab/exemplary-empathy}{https://github.com/declare-lab/exemplary-empathy}}

\noindent \textbf{ChatGPT+Causality} \cite{fu2023reasoning}: which is based on a commonsense-based causality explanation that considers both the user's and the system's perspective to enhance ChatGPT's ability for empathetic response generation.

\subsubsection{Ablation Studies in Proposed StyEmp}
We utilize DialoGPT as the base decoder across all ablation studies. The proposed StyEmp model integrates a multi-grained prefix encoder (MgPE (C+E+P)) with personality reinforcement in the decoder (DialoGPT w/ PR). To explore the efficacy of each component within the encoder and decoder, we conduct ablation studies using four configurations of the multi-grained prefix encoder:
\noindent(1) \textbf{MgPE (C+E+P)}:  includes both the context-personality-aware prefix encoding and context-empathy-aware prefix encoding. 
In addition, there are other three configurations:
(2) \textbf{MgPE (C)} incorporates only context-aware prefix encoding;
(3) \textbf{MgPE (C+P)} includes only context-personality-aware prefix encoding;
(4) \textbf{MgPE (C+E)} integrates only context-empathy-aware prefix encoding.

These are evaluated under two conditions in the decoder: \textbf{DialoGPT w/ PR} (with PR integration) and \textbf{DialoGPT w/o PR} (without PR integration). 

\begin{table*}[t]
\centering
\scalebox{0.85}{
\begin{tabular}{@{}lrr|rr|ll|lll@{}}
\toprule
\multirow{2}{*}{Methods} & \multicolumn{2}{c}{Semantics} & \multicolumn{2}{c}{Diversity} & \multicolumn{2}{c}{Personality} & \multicolumn{3}{c}{Empathy}     \\ \cmidrule(l){2-10} 
                         & BERTS       & BLEURT      & D1            & D2            & E\&I         & T       & EAcc. & IP\&EX & Intent \\ \midrule
\multicolumn{10}{@{}l@{}}{\textit{Transformer-based methods}} \\ 
MOEL                     & 52.67          & 34.48      & 0.44          & 2.02          & 0.0525          & 0.0525        & 26.80   & 70.06 & 22.77       \\
MIME                     & 52.87          & 35.64      & 0.32          & 1.12          & 0.0200         & 0.0675        & 22.40   & 70.17  & 25.11       \\
EmpDG                    & 51.99          & 34.60        & 0.79          & 3.23          & 0.0155          & 0.1115        & 26.49   & 68.09  & 21.29       \\
CEM                      & 52.41           & 35.06       & 0.65          & 2.92          & 0.0741         & 0.1519      & 32.85   & \textbf{73.62}  & 29.37       \\ \midrule
\multicolumn{10}{@{}l@{}}{\textit{Large language model-based methods}}       \\ 
LEMPEx                   & 49.03           & 27.92       & 1.20           & 12.88         & -0.0077         & 0.0706        & 31.73     & 69.03  & 27.99       \\
DialoGPT                 & \underline{54.24}           & 40.32       & \underline{2.92}          & 15.62        & 0.1361          & 0.1723        & 33.68   & 72.49 & 31.53       \\
ChatGPT+Causality        & \textbf{54.93}           & \textbf{43.45}       & 2.91          & \textbf{16.44}         & 0.1584          & 0.1774        & 30.79    & 69.64   & 27.86        \\ 
\midrule
\multicolumn{10}{@{}l@{}}{\textit{Our proposed method}}   \\

StyEmp w/o PR           & 54.13           & \underline{41.00}           & \textbf{2.95}          & \underline{16.10}         & \underline{0.1681}          & \underline{0.2010}        & \underline{34.47}    & 72.70  & \underline{31.73}        \\ 
StyEmp &53.60	&40.49	&2.21	&9.48    & \textbf{0.1758$^{*}$}         & \textbf{0.2093$^{*}$}       &  \textbf{34.88$^{*}$} & \underline{73.02$^{*}$} & \textbf{31.85$^{*}$} \\
\bottomrule
\end{tabular}}
\caption{Objective evaluation results of baselines and our proposed method. \textbf{Bold} and \underline{underline} denote the best and second-best score, respectively. $^{*}$ indicates a statistically significant difference for $p<0.05$ between StyEmp and ChatGPT+Causality, determined by t-test.}
\label{tab:autmatic}
\end{table*}

\begin{table*}[t]
\centering
\scalebox{0.89}{
\begin{tabular}{@{}lrr|rr|rr|rrr@{}}
\toprule
\multirow{2}{*}{Methods} & \multicolumn{2}{c}{Semantics} & \multicolumn{2}{c}{Diversity} & \multicolumn{2}{c}{Personality} & \multicolumn{3}{c}{Empathy}     \\ \cmidrule(l){2-10} 
                         & BERTS       & BLEURT      & D1            & D2            & E\&I         & T       & EAcc. & IP\&EX & Intent \\ \midrule
DialoGPT w/o PR        & 54.24           & 40.32       & 2.92          & 15.62        & 0.1361          & 0.1723        & 33.68   & 72.49 & 31.53       \\
+MgPE (C)     & \underline{54.43}           & \underline{41.18}       & 2.85          & 16.08    & 0.1525          & 0.1828        & 34.08   & 72.57  & 31.00          \\
+MgPE (C+P)  & 53.99           & 40.31       & \textbf{3.07}          & \textbf{16.80}     & 0.1639         & 0.1987        & 34.30     & 71.71 & 31.47       \\
+MgPE (C+E) & \textbf{54.55}           & \textbf{41.25}       & 2.87          & 15.80      & 0.1552        & 0.1890         & 34.32    & 72.90 & 31.75       \\

+MgPE (C+E+P) & 54.13           & 41.00         & \underline{2.95}          & \underline{16.10}         & 0.1681          & 0.2010        & 34.47    & 72.70  & 31.73        \\ 
\midrule

DialoGPT w/ PR    &53.92	&40.37	&2.23	&9.74 & 0.1672          & 0.1824        & 34.37    & \underline{73.42}  & \bf{32.23}       \\ 

+MgPE (C)   &53.96	&40.83	&2.22	&9.63 & 0.1669          & 0.1997        & \underline{35.37}    & 72.76  & 31.14       \\ 

+MgPE (C+P)  &53.24	&40.29	&2.05	&8.93
                 &   \underline{0.1683} &    \bf{0.2108}             & 34.14              & 72.81         &  31.42                  \\ 

+MgPE (C+E) &53.89	&40.52	&2.32	&9.89 & 0.1680                 & 0.1949              &  \bf{35.65}        &  \bf{73.58}      &   \underline{32.21}          \\ 

+MgPE (C+E+P) &53.60	&40.49	&2.21	&9.48    & \textbf{0.1758}         & \underline{0.2093}       &  34.88 &73.02 &31.85  
\\ \bottomrule
\end{tabular}}
\caption{Ablation studies on the effect of context, past responses (implicit personality), empathy explanation in the multi-grained prefix encoder, and explicit personality reinforcement (PR) module.}
\label{tab:ablation}
\end{table*}

\subsection{Evaluation Metrics}
\subsubsection{Objective Evaluations}

\noindent \textbf{BERTScore} \cite{zhang2019bertscore}: a BERT-based evaluation metric, which focuses on lexical semantic similarity between the
generated response and the ground truth. We adopt
its F1 score and use the "deberta-large-mnli" version.\footnote{\href{https://github.com/Tiiiger/bert\_score}{https://github.com/Tiiiger/bert\_score}}

\noindent \textbf{BLEURT} \cite{sellam-etal-2020-bleurt}: evaluates to what extent the generated response is fluent and conveys the meaning of the reference.\footnote{\href{https://github.com/google-research/bleurt}{https://github.com/google-research/bleurt}}

\noindent \textbf{D1/D2} (Distinct-1/2) \cite{li2016diversity}: counts the number of distinct n-grams in generated responses.

\noindent \textbf{E\&I}: denotes the mean Pearson correlation coefficient between the ground truth and generated responses for extroversion (E) from the Big 5 predictor and introversion (I) from the MBTI predictor.

\noindent \textbf{T}: represents the Pearson correlation coefficient between the ground truth and generated responses for thinking (T) from the MBTI predictor.

\noindent \textbf{EAcc.}: refers to the average accuracy of both emotion (Emo.) and ER prediction, comparing the generated responses with ground truth. 

\noindent \textbf{IP\&EX}: refers to the average accuracy of both interpretation (IP) and exploration (EX) prediction, comparing generated responses with ground truth.

\noindent \textbf{Intent}: accuracy of empathetic intent prediction between the generated responses and ground truth.

\subsubsection{Human Evaluations}
We randomly select 100 samples from the test set across all models. Each sample is evaluated by three different crowd-workers hired through Amazon Mechanical Turk. More details can be seen in Appendix~\ref{sec:HIT}.
We assess the quality of these responses based on two criteria, each criterion is rated on a 1 to 5 scale: 
(1) \textbf{Empathy}, determining if the generated responses demonstrate understanding of the speaker's feelings and experiences. (2) \textbf{Personality}, refers to personality consistency; we provide crowd-workers with five sampled past responses from the listener of the ground truth and ask them to evaluate if the generated response aligns with the listener's personality traits.

\section{Results and Analysis}
\subsection{Objective Evaluation Results}
Table~\ref{tab:autmatic} presents the automatic evaluation results for both baselines (including transformer-based and LLM-based methods), and our proposed method. The results illustrate that our method significantly outperforms the baselines in terms of personality, emotion, and intent accuracy, while maintaining the semantic scores comparable to DialoGPT.  
The proposed StyEmp with PR degrades the semantic score because it re-ranks the original output of DialoGPT by weighting the personality consistency. 


We also conducted ablation studies to evaluate different encoder configurations, comparing their performance in scenarios with and without PR. As depicted in Table~\ref{tab:ablation}, 
In both scenarios, MgPE (C+P) and MgPE (C+E) surpass MgPE (C) on most personality and empathy metrics. Moreover, MgPE (C+P+E) further outperforms both MgPE (C+P) and MgPE (C+E). These results support our hypothesis that empathy and personality enrich each other. Incorporating PR further enhances the expression of both traits. These findings show the substantial contribution of the PR module in enhancing model performance for generating responses that are both empathetic and reflective of distinct personalities.

\subsection{Human Evaluation Results}
Table~\ref{tab:human} shows that our methods rank highest against baselines. Specifically, DialoGPT with the proposed MgPE (C+E+P) and MgPE (C+E+P) w/ PR significantly outperform finetuned DialoGPT, enhancing empathy and personality expression in generated responses. However, StyEmp performs worse than MgPE (C+E) w/ PR and MgPE (C+E+P) w/o PR regarding personality, inconsistent with the objective evaluation results. This discrepancy stems from inaccuracies in personality prediction, particularly when conflicts arise between the predicted personality traits and those implied by past responses. This is a limitation of using personality predictor with accuracy of 60-70\%. More error analysis can be found in Appendix~\ref{sec:cases}. 
\begin{table}[h]
\centering
\label{tab:my-table}
\begin{tabular}{@{}lll@{}}
\toprule
Models            & Empathy & Personality   
\\ \midrule
CEM               & 3.35    & 2.93  
\\
ChatGPT+Causality & 4.00    & 3.11  
\\\midrule
DialoGPT          & 3.04    & 2.99   
\\
+MgPE (C+E+P)     & \underline{4.05}$^{*}$    &  \underline{3.25}$^{*}$        
\\
+MgPE (C+E) w/ PR     &  3.97             &  \textbf{3.39}  
\\
+MgPE (C+E+P) w/ PR           & \textbf{4.08}$^{*}$    & 3.18$^{*}$   
\\ 
\bottomrule
\end{tabular}
\caption{Results of human evaluations. DialoGPT+ MgPE (C+E+P) w/ PR refers to StyEmp. $^{*}$ indicates a statistically significant improvement ($p<0.05$) over DialoGPT.}
\label{tab:human}
\end{table}

\subsection{Case Studies}
Table~\ref{tab:casestudy} compares our proposed StyEmp model with baseline methods, highlighting differences in personality trait expression. The baseline methods fall short of showing explicit personality traits, often resulting in more general responses. On the other hand,  StyEmp showcases extroverted traits (predicted by our method), utilizing expressions like \textit{"wow, bet"} and longer phrases in this example. Moreover, the StyEmp-generated responses are more closely aligned with the personality traits shown in the ground truth, indicating its effectiveness in accurately reflecting personality. More examples are shown in the Appendix~\ref{sec:cases}.
\begin{table}[htp]
\centering
\scalebox{0.63}{
\begin{tabular}{@{}ll@{}}
\toprule
Context          & \begin{tabular}[c]{@{}l@{}}
I studied so hard for 3 months straight for my bar exam\\ to become a lawyer.
\end{tabular}      \\
Ground truth     &\begin{tabular}[c]{@{}l@{}} \underline{wow}, you're so determined! Did you pass your exam? \end{tabular}\\\midrule
MoEL              & That is awesome! I hope you do well!      \\
MIME              & That is great. I am sure you will do great! \\
EmpDG             & That is great! What did you do?    \\
CEM               & that is great! I am sure you will do great! \\
LEMPEx            & \begin{tabular}[c]{@{}l@{}}Congratulations! That’s awesome! Congratulations. \end{tabular} \\
DialoGPT          & That’s great, I hope you did well. \\
\begin{tabular}[c]{@{}l@{}}ChatGPT+\\Causality\end{tabular} &  \begin{tabular}[c]{@{}l@{}} Congratulations on all your hard work and dedication! \end{tabular} \\ \midrule
\multicolumn{2}{@{}l@{}}{\textit{Predicted system's \colorbox{yellow!50}{personality: Extrovert, Feeling}}}  \\
\multicolumn{2}{@{}l@{}}
{\textit{Predicted system's \colorbox{antiquewhite}{Empathy: Emotional reaction; Emotion intent is wishing.}}} \\ 
\begin{tabular}[c]{@{}l@{}}StyEmp \\ w/o PR \end{tabular}   & \begin{tabular}[c]{@{}l@{}}That's great! That's the best feeling in the world!  \\ What are you studying? \end{tabular}                                             \\
StyEmp & \begin{tabular}[c]{@{}l@{}}\underline{Wow}, that's a long time! I \underline{bet} you were really proud of \\yourself! What kind of bar did you study? I hope you did well!\end{tabular} \\ \bottomrule
\end{tabular}}
\caption{Comparative case studies between our proposed StyEmp and baselines.}  
\label{tab:casestudy}
\end{table}

\section{Conclusions and Future Work}
We have proposed StyEmp, which aims to stylize empathetic response generation with consistent personality. Specifically, StyEmp incorporates a multi-grained prefix mechanism designed to capture the intricate relationship between a system's personality and its empathetic expressions. Furthermore, we introduce a personality reinforcement module that leverages contrastive learning to calibrate the generation model, ensuring responses are both empathetic and reflective of the distinct personality. The experimental results demonstrate that our method outperforms other competitive methods on both automatic and human evaluations.

The performance of our model is currently limited by the efficacy of the personality predictor.
In future work, we plan to utilize ground-truth personality traits instead of predicted ones by annotating the dataset with personality labels.

\section*{Limitations}
Given our objective to enrich responses with empathy and personality information, we face the challenge of a scarcity of datasets that provide both empathety and personality annotations. 
Therefore, we have developed additional personality scorers, as shown in Table~\ref{tab:personalityacc} and detailed in Appendix~\ref{sec:personality_anno}. However, the results from these scorers are not ideal, significantly impacting the effectiveness of our personality reinforcement module, since we rely on the predicted personality to enhance the system's personality expression. To overcome this limitation, we plan to collect a dataset that includes both empathy and personality annotations in future work.

\section*{Acknowledgements}
This work was supported by JST Moonshot R\&D Goal 1 Avatar Symbiotic Society Project (JPMJMS2011). This work was also supported by JST SPRING, Grant Number JPMJSP2110.

\bibliography{anthology,custom}
\bibstyle{acl_natbib}

\appendix

\section{Personality Predictor}
\label{sec:personality_anno}
We implemented strict speaker splitting to ensure no overlap among speakers across the training, validation, and test sets. This approach ensured that the model was evaluated on unseen speakers, thereby making the evaluation results on the PANDORA dataset applicable to the ED dataset as well. The Big 5 personality trait scores are continuous, ranging from -100 to 100, while MBTI scores are binary. 
We normalized each Big 5 personality trait score to a range between -1 and 1 and balanced the binary labels of each MBTI trait, The details of the statistics are shown in Table~\ref{tab:statistic_personality} for reference. 

To make the length distribution of the examples similar to the ED dataset, we conducted the following steps for both Big 5 and MBTI experiments: 1) only preserved sentences containing ASCII characters with 10 to 50 tokens. 2) For each user we derived non-overlapping samples by randomly selecting and concatenating \textit{k} sentences, where \textit{k} was randomly selected to vary between 1 and 5. 

We incorporated five fully connected layers with ReLU activation followed by five regression heads on top of the LUKE model, to predict all Big 5 trait intensities simultaneously.
We separately finetune the LUKE model with one fully connected layer and one regression head for each MBTI trait prediction. For all the experiments, the learning rate is set as 1e-5, the dropout is 0.1, and the mean squared error loss. We used a linear scheduler with a warmup step of 100. Using the median of the training label and 0.5 as the threshold, we further binarize the predicted intensities and actual labels and report the accuracies and F1 scores for Big 5 and MBTI, separately.

\begin{table*}[t]
\centering
\scalebox{0.95}{
\begin{tabular}{@{}llcccc@{}}
\toprule
                      & Traits                       & unique                         & train             & valid             & test              \\ \midrule
\multirow{8}{*}{MBTI} & \multirow{2}{*}{Introverted} & speakers                       & 1,531 | 1,402       & 197 | 170         & 193 | 174         \\
                      &                              & utterances                     & 412,467 | 424,008 & 55,870 | 48,218   & 49,167 | 56,177   \\
                      & \multirow{2}{*}{Intuitive}   & speakers                       & 820 | 995         & 100 | 126         & 106 | 120         \\
                      &                              & utterances                     & 268,470 | 277,440 & 38,443 | 30,230   & 34,022 | 34,527   \\
                      & \multirow{2}{*}{Thinking}    & speakers                       & 2,568 | 1,728       & 307 | 230         & 334 | 205         \\
                      &                              & utterances                     & 547,753 | 561,814 & 70,483 | 66,916   & 72,527 | 66,181   \\
                      & \multirow{2}{*}{Perceiving}  & speakers                       & 2,965 | 3,110       & 388 | 371         & 392 | 367         \\
                      &                              & utterances                     & 871,439 | 877,865 & 109,267 | 108,546 & 107,740 | 112,082 \\ \cmidrule(l){2-6} 
\multirow{2}{*}{Big5} & \multirow{2}{*}{All}         & \multicolumn{1}{l}{speakers}   & 1,225              & 153               & 154               \\
                      &                              & \multicolumn{1}{l}{utterances} & 102,523           & 12,803            & 12,803            \\ \bottomrule
    \end{tabular}}
\caption{Statistics of unique speakers and utterances across each MBTI and all Big 5 traits in the filtered Pandora dataset. For MBTI traits, we show the number of label 0 | 1.}
\label{tab:statistic_personality}
\end{table*}

\section{Case Studies}
\label{sec:cases}
We present two examples comparing our StyEmp model with baseline methods, showcasing StyEmp's superior ability to convey consistent personality traits, as shown in Table~\ref{tab:morecases}.

We further show two examples that our StyEmp failed to show consistent personality because of incorrect personality prediction. In contrast, StyEmp without PR correctly expresses personality by learning from past responses by the same listener from the training set, as shown in Table~\ref{tab:error}.

\section{Human Evaluation Instruction}
\label{sec:HIT}
\begin{figure*}
    \centering
\includegraphics[width=1\textwidth]{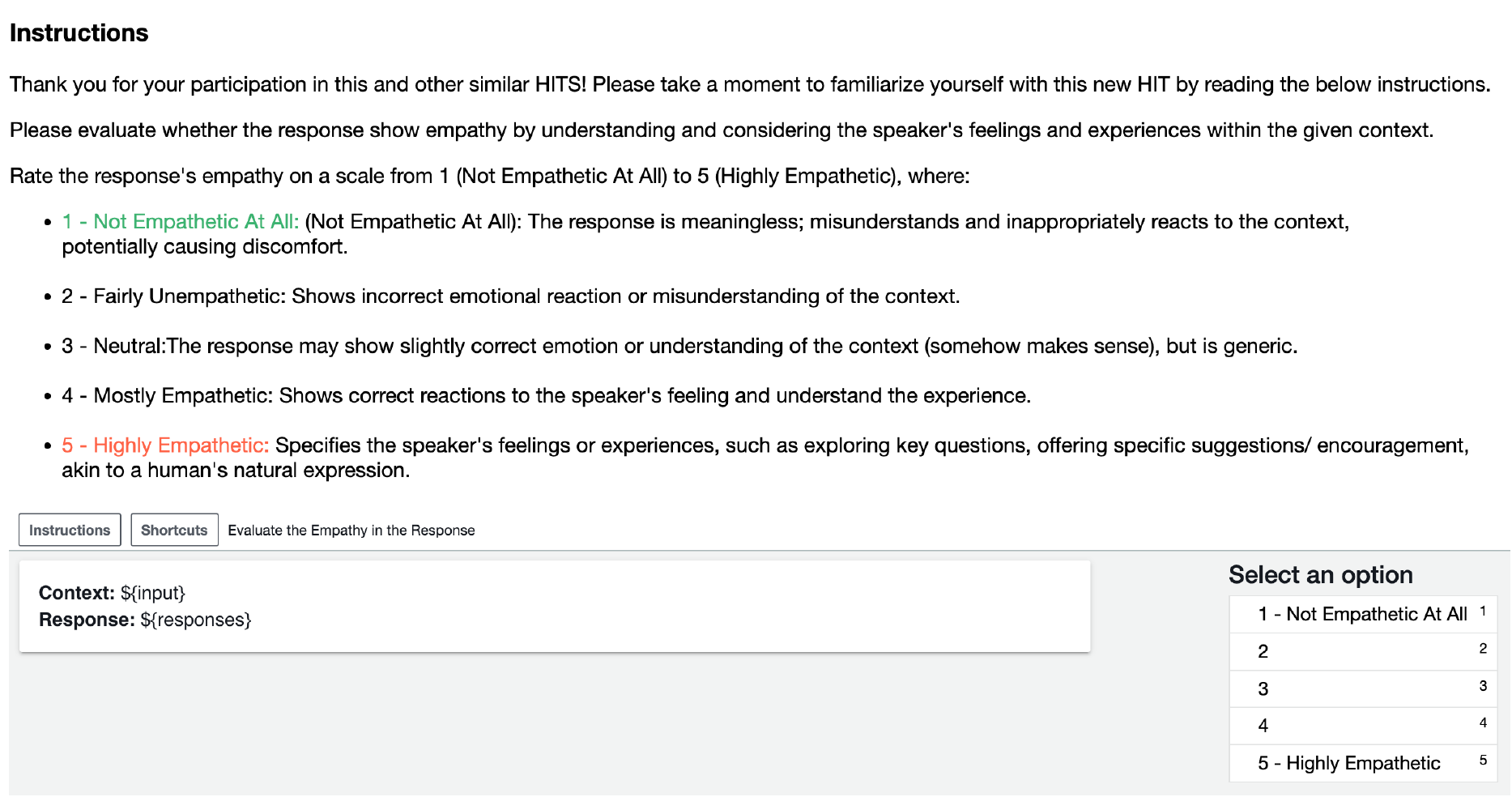}
    \caption{Template for human evaluation on empathy in generated responses.}
    \label{fig:hit_emp}
\end{figure*}

\begin{figure*}
    \centering
\includegraphics[width=1\textwidth]{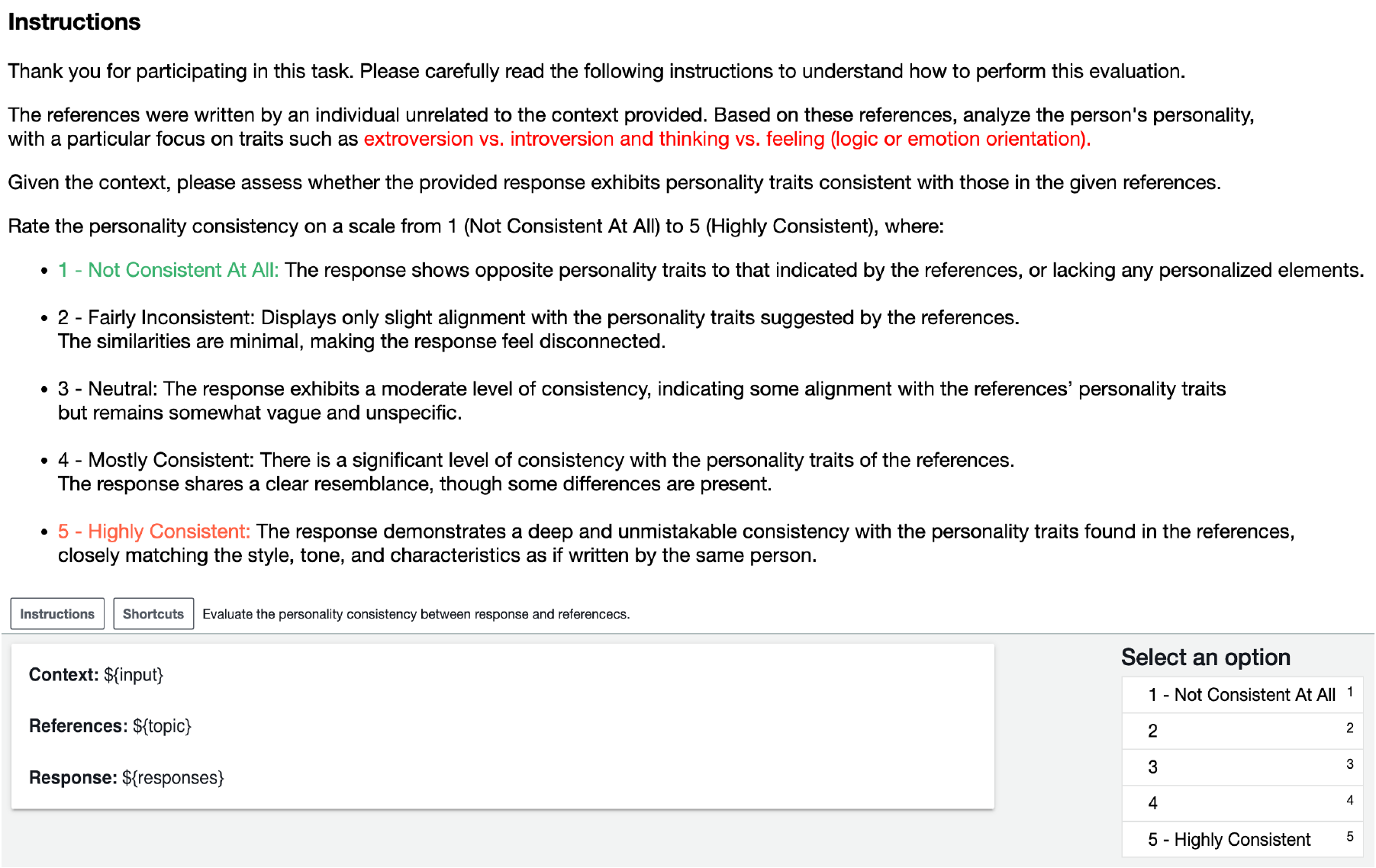}
    \caption{Template for human evaluation on personality consistency in generated responses.}
    \label{fig:hit_per}
\end{figure*}

Human evaluation is necessary for an open-domain dialogue system. 
We randomly select 100 context-response pairs from the test set across all models. Each response is evaluated by three different crowd-workers, provided with the corresponding context. We hired crowd workers through Amazon Mechanical Turk, and each has a historical approval rate of over 98\% on human evaluation tasks. We assess the quality of generated responses by our proposed method and comparative models based on empathy and personality consistency, the template for the human evaluations is shown in Figure~\ref{fig:hit_emp} and Figure~\ref{fig:hit_per}.

\begin{table*}[t]
\scalebox{0.9}{
\begin{tabular}{@{}lll@{}}
\toprule                                
Context          & \begin{tabular}[c]{@{}l@{}}I felt so nervous going to my last job interview!\\ Nerves must occur in proportion to how badly you *want* something!
 \end{tabular}           
                    \\
Ground truth     & I know what you mean. Did you get the job?\\
\midrule
MoEL              &  I am sorry to hear that. I hope you do well!          \\
MIME              & I am sure you will do well.       \\
EmpDG             & I am sure you will be fine.    \\
CEM               & I am sure you will be fine.    \\
LEMPEx            & That's a great job, I am sure you will do well.     \\           
DialoGPT          & I'm sure you'll do great! I'm sure you'll do great! \\
ChatGPT+Causality & \begin{tabular}[c]{@{}l@{}}I understand how you feel, but remember to take deep breaths and focus on your strengths.\\ You got this! \end{tabular}    \\ \midrule
\multicolumn{2}{@{}l@{}}{\textit{Predicted system's \colorbox{yellow!50}{personality: introvert, feeling}}}  \\
\multicolumn{2}{@{}l@{}}{\textit{Predicted system's \colorbox{antiquewhite}{Empathy: Emotional reaction; Emotion intent is acknowledging.}}}\\ 
StyEmp w/o PR       & \begin{tabular}[c]{@{}l@{}}I feel the same way. I hope you get the job!\end{tabular}    \\
StyEmp  & \begin{tabular}[c]{@{}l@{}}I know what you mean. I'm nervous too.\end{tabular} 
                     \\ \midrule \midrule
                     
Context           & \begin{tabular}[c]{@{}l@{}} I went back to my home where I grew up and a flood of good memories came back to me.\\ I had so many fun times with my best friends when I was a child. \end{tabular}  
                    \\
Ground Truth      & What is your favorite memory with your friends?
     \\
\midrule
MoEL              & That is so sad. I bet you were a great time.   \\
MIME              & That is so sweet. I am glad you had a good time.  \\
EmpDG             & I bet you were so proud of them.  \\
CEM               & That is awesome. I bet you were very happy.
    \\
LEMPEx            & That must have been a cool old times!  \\
DialoGPT          & I bet you were so happy to have that feeling.
 \\
ChatGPT+Causality & \begin{tabular}[c]{@{}l@{}}That sounds amazing! I love reminiscing about my childhood too.\\ Do you have any specific memories that stand out to you?\end{tabular}              
                   \\ \midrule
\multicolumn{2}{@{}l@{}}{\textit{Predicted system's \colorbox{yellow!50}{personality: extrovert, feeling}}}  \\
\multicolumn{2}{@{}l@{}}{\textit{Predicted system's \colorbox{antiquewhite}{Empathy: Interpretation. Emotion Intent is neutral.}}}\\
StyEmp w/o PR       & \begin{tabular}[c]{@{}l@{}}I can imagine!  It was nice to have good memories that you can cherish as a child.\\  I'm glad you had good memories and not too many bad memories.\end{tabular}     \\
StyEmp  & \underline{Oh wow}, I bet that was a great time. I bet you were so happy to have them back.  \\

\bottomrule
\end{tabular}
}
\caption{Comparative case studies between baselines and our proposed StyEmp.}
\label{tab:morecases}
\end{table*}

\begin{table*}[h]
\scalebox{0.89}{
\begin{tabular}{@{}lll@{}}
\toprule                                         
Context          & \begin{tabular}[c]{@{}l@{}}I one time lost my trunks in the pool! People saw me in a way I didn't want!
\end{tabular}           
                    \\
Ground truth     & \begin{tabular}[c]{@{}l@{}}\underline{Oh no}! That must have been super embarrassing! How did you react to that? \end{tabular}\\
\midrule
\multicolumn{2}{@{}l@{}}{\textit{Predicted system's \colorbox{gray!30}{personality: introvert, feeling}}}   
\\ 
\multicolumn{2}{@{}l@{}}{\textit{Predicted system's \colorbox{antiquewhite}{Empathy: Emotional Reaction, Exploration. Emotion Intent is questioning.}}}\\
StyEmp w/o PR       & \begin{tabular}[c]{@{}l@{}}\underline{Oh no}! Did you get them back?
\end{tabular}  
 \\
StyEmp  & \begin{tabular}[c]{@{}l@{}}Did you get hurt?
\end{tabular}\\  \midrule
Context & \begin{tabular}[c]{@{}l@{}}My friend came over yesterday and we were sitting on the couch chit-chatting and when I got up\\ I accidentally farted. So embarrassing.\end{tabular}\\
Ground truth & \underline{Oh my}, did they notice you farted? \\ \midrule
\multicolumn{2}{@{}l@{}}{\textit{Predicted system's \colorbox{gray!30}{personality: introvert, feeling}}}  
\\ 
\multicolumn{2}{@{}l@{}}{\textit{Predicted system's \colorbox{antiquewhite}{Empathy: Exploration. Emotion Intent is questioning.}}}\\
StyEmp w/o PR & \underline{Oh no!} Did you say anything to him?\\
StyEmp     & Did you apologize?\\
\bottomrule
\end{tabular}
}\caption{Cases exist where StyEmp fails to accurately express the intended personality due to \colorbox{gray!30}{errors} in personality prediction, which lead to errors in PR. In contrast, StyEmp without PR correctly expresses personality by learning from past responses by the same listener from the training set.}
\label{tab:error}
\end{table*}


\clearpage

\end{document}